\documentclass[sigconf]{acmart}
\AtBeginDocument{%
  \providecommand\BibTeX{{%
    \normalfont B\kern-0.5em{\scshape i\kern-0.25em b}\kern-0.8em\TeX}}}

\copyrightyear{2023}
\acmYear{2023}
\setcopyright{none}
\acmConference[CIKM '23]{Proceedings of the 32nd ACM International Conference on Information and Knowledge Management}{October 21--25, 2023}{Birmingham, United Kingdom}
\acmBooktitle{Proceedings of the 32nd ACM International Conference on Information and Knowledge Management (CIKM '23), October 21--25, 2023, Birmingham, United Kingdom}
\acmPrice{15.00}
\acmDOI{XXXXXXX.XXXXXXX}
\acmISBN{978-1-4503-XXXX-X/18/06}

\usepackage{subcaption}
\usepackage{balance}

\begin{document}

\title{Time-aware Graph Structure Learning via Sequence Prediction on Temporal Graphs}

\author{Haozhen Zhang}
\authornote{Work was done during internship at Microsoft Research Asia.}
\affiliation{%
  \institution{Shenzhen International Graduate School}
  \city{Tsinghua University, Shenzhen}
  \country{China}
  }
\email{zhang-hz21@mails.tsinghua.edu.cn}

\author{Xueting Han}
\authornote{Corresponding author.}
\affiliation{%
  \institution{Microsoft Research Asia}
  \city{Beijing}
  \country{China}
  }
\email{chrihan@microsoft.com}

\author{Xi Xiao}
\affiliation{%
  \institution{Shenzhen International Graduate School}
  \city{Tsinghua University, Shenzhen}
  \country{China}
  }
\email{xiaox@sz.tsinghua.edu.cn}

\author{Jing Bai}
\affiliation{%
  \institution{Microsoft Research Asia}
  \city{Beijing}
  \country{China}
  }
\email{jbai@microsoft.com}

\begin{abstract}

Temporal Graph Learning, which aims to model the time-evolving nature of graphs, has gained increasing attention and achieved remarkable performance recently. 
However, in reality, graph structures are often incomplete and noisy, which hinders temporal graph networks (TGNs) from learning informative representations.
Graph contrastive learning uses data augmentation to generate plausible variations of existing data and learn robust representations. However, rule-based augmentation approaches may be suboptimal as they lack learnability and fail to leverage rich information from downstream tasks.
To address these issues, we propose a \textbf{T}ime-aware \textbf{G}raph \textbf{S}tructure \textbf{L}earning (\textbf{TGSL}) approach via sequence prediction on temporal graphs, which learns better graph structures for downstream tasks through adding potential temporal edges. In particular, it predicts time-aware context embedding based on previously observed interactions and uses the Gumble-Top-K to select the closest candidate edges to this context embedding. Additionally, several candidate sampling strategies are proposed to ensure both efficiency and diversity. Furthermore, we jointly learn the graph structure and TGNs in an end-to-end manner and perform inference on the refined graph. Extensive experiments on temporal link prediction benchmarks demonstrate that TGSL yields significant gains for the popular TGNs such as TGAT and GraphMixer, and it outperforms other contrastive learning methods on temporal graphs. We release the code at \url{https://github.com/ViktorAxelsen/TGSL}.

\end{abstract}




\maketitle

\section{Introduction}
\label{sec:intro}

Graph representation learning has been found powerful in addressing complex tasks involving graphical data such as social networks, interaction networks, and biomolecular graphs recently~\cite{GNN_Application}. However, most approaches utilize static graph neural networks (GNNs)~\cite{GCN, GraphSAGE, GAT} and fail to consider the time-evolving nature of some graph data, called dynamic or temporal graphs~\cite{Dynamic_Graph_Survey}, resulting in poor performance. To address this issue, recent research has proposed temporal graph networks (TGNs)~\cite{CAW, TGAT, TGN}. Some of these methods transform the graph into a sequence of discrete snapshots, known as discrete-time dynamic graphs (DTDG)~\cite{DTDG}, and process them separately using static GNNs. Conversely, others directly process the interactions that occur in time order on the graph, known as continuous-time dynamic graphs (CTDG)~\cite{TGN}. While the former may lose some time information due to the discrete nature of the processing, the latter exhibit finer time granularity and typically achieves better results. Therefore, we primarily focus on CTDG in this paper.

Despite the success of GNNs, their effectiveness is heavily dependent on the quality of the data. 
However, in real-world scenarios, graph structures may contain flaws due to oversights during collection or processing, including issues such as incomplete, noisy, and redundant links or nodes ~\cite{GSL_Survey}. 
For example, in interaction graphs, it is common to lose interactions between entities due to unavoidable factors, resulting in incomplete graphs~\cite{GSR}. 
Since GNNs compute node embeddings by recursively aggregating information from neighbors, the prevalent message-passing design is negatively impacted by flawed graph structures. When using such flawed graphs, the learned graph representations will also inherit the flaws present in the original graph structure, causing a bottleneck in improving the model performance.

Recently, with the rise of self-supervised learning (SSL) and its general application in the field of graph representation learning~\cite{contrastive_multiview_graph, DGI, GCC, GraphCL, GCA, GraphMAE}, there are some methods utilizing SSL, particularly graph contrastive learning (GCL)~\cite{GraphCL, GCC, GCA}, to alleviate the challenges mentioned above and improve the performance on temporal graphs. 
GCL creates plausible variations of existing data without the need for additional labeled data and employs a self-supervised objective to learn representations that are resilient to noise and perturbations by maximizing the agreement between learned representations. In the context of temporal graphs, GCL approaches employ time-dependent graph data augmentation (GDA) to construct more powerful views of the temporal graph with different graph structures or features and then contrast two views by utilizing contrastive loss, such as InfoNCE~\cite{CPC}, to encourage the model to learn more robust and effective representations~\cite{MeTA, DDGCL}. These techniques enable models to better generalize across those variations and attend to signal over noise. 
However, these methods construct graph views based on rule-based perturbations of the original graph. Since the augmented structures are non-learnable and unable to leverage the rich information from downstream tasks, they cannot optimize or mitigate the flaws in the original graph. As a result, they may not necessarily generate improved graph structures for downstream tasks, leading to suboptimal performance.

To address the above problems, graph structure learning (GSL) has been proposed in recent years, which is a very promising approach to improve the performance of GNN from the perspective of refining the original graph structure~\cite{SLAPS, SUBLIME, GRCN, ProGNN, IDGL}. 
Essentially, GSL treats the graph structure as learnable parameters and iteratively optimizes them while learning the model parameters. Generally, GSL leverages node features to reconstruct graph structure, followed by post-processing operations. 
The learned graph structure is often superior to the original one and can be used for further training and inference. 
However, most current graph structure learning methods are designed for static graphs and do not consider time-related information. 
Moreover, they rely heavily on the initial node features, which are lacking in temporal graphs~\cite{GSL_Survey}. 
Consequently, current GSL methods fail to exploit time-related information from interactions on temporal graphs, leading to inadequate graph structure construction.

Based on the above observations, in this paper, we propose a novel \textbf{T}ime-aware \textbf{G}raph \textbf{S}tructure \textbf{L}earning (\textbf{TGSL}) approach via sequence prediction to refine graph structure on temporal graphs. 
TGSL can be applied to existing TGNs such as TGAT~\cite{TGAT} and GraphMixer~\cite{GraphMixer}. 
TGSL learns to add potential edges on temporal graphs through sequence prediction, and the learned graph structure is used with the original one in a contrastive way during training. Specifically, our contributions include: 
\begin{itemize}
\item TGSL first employs an edge-centric time-aware graph neural network (ET-GNN) to extract edge embeddings, and the previously interacted neighbors of each node are sequentially fed into a recurrent neural network to predict the time-aware context embedding, which represents the neighborhood information of a node at a certain timestamp.
\item In candidate edge construction, instead of calculating on the full graph, we propose several candidate sampling strategies to improve efficiency and enhance the diversity of candidate edges. Additionally, a time-mapping mechanism is employed to project the context embedding and candidate edge embeddings to the newly sampled timestamps. The weights of candidate edges are then computed based on the similarity between the projected context embedding and candidate embeddings. Inspired by reparameterization tricks, We use the Gumble-Top-K trick~\cite{Gumble-top-k} to select $K$ edges for addition, further facilitating the exploration of edge diversity.
\item We use multi-task learning to jointly optimize TGNs and the proposed TGSL in an end-to-end manner, and we directly utilize the refined graph instead of the original one for inference. 
\item Extensive experiments demonstrate that TGSL remarkably improves performance. For example, it improves transductive accuracy and average precision of TGAT by 3.3\% and 1.3\%, and GraphMixer by 1.2\% and 0.6\% on the Wikipedia dataset w.r.t. temporal link prediction. 
Further elaborately designed experiments also show TGSL outperforms other contrastive learning methods on temporal graphs. 
\end{itemize}

\section{Related Work}
\label{sec:related_work}

\subsection{Temporal Graph Networks}

Most GNNs are designed for static graphs, while temporal graphs are constantly evolving over time. 
Therefore, temporal graph networks (TGNs) have been proposed to tackle the above problem~\cite{Dynamic_Graph_Survey}. 
Basically, there are two kinds of temporal graphs: discrete-time dynamic graphs (DTDG)~\cite{DTDG} and continuous-time dynamic graphs (CTDG)~\cite{TGN}. 
Early works mainly focus on the DTDG, which represents a temporal graph as a sequence of discrete snapshots. They usually employ a static GNN, such as GAT~\cite{GAT}, GraphSAGE~\cite{GraphSAGE}, etc., to encode snapshots separately, and an aggregation strategy is utilized to fuse the outputs of GNN for final prediction~\cite{ROLAND, EvolveGCN, DySAT, DynGEM}. 
However, some time-continuous information is inevitably lost due to discrete processing, resulting in model performance degradation. 
CTDG addresses this defect in DTDG by representing the temporal graph as a time-continuous event stream and usually can achieve better results. 
For example, DyRep~\cite{DyRep} divides the time-evolving process of temporal graphs into association and communication processes and updates node embeddings via temporally attentive aggregation. 
JODIE~\cite{JODIE} employs RNNs to update user and item embeddings, respectively and uses an embedding projection operation to predict future embeddings. 
TGAT~\cite{TGAT} directly aggregates the central node's neighbors before event timestamps in a self-attentive manner, and CAW~\cite{CAW} samples temporal random walks to encode temporal graph dynamics. 
TGN~\cite{TGN} further enhances TGAT~\cite{TGAT} with node memories and achieves better performance and training efficiency. 
Recently, GraphMixer~\cite{GraphMixer} simplifies the design of TGNs and utilizes MLP-mixer~\cite{MLP-MIXER} for link encoding.

\subsection{Graph Contrastive Learning}
Graph contrastive learning (GCL) is a widely used graph self-supervised learning method due to its intuitiveness and effectiveness~\cite{DGI, GCC, GCA, GraphCL}. 
Take GraphCL~\cite{GraphCL} as an example, two graph views are generated by different graph data augmentations such as edge perturbation, feature masking, etc., and a shared GNN encoder is employed to encode these two graph views. 
Specifically, GraphCL uses the NT-Xcent loss function, which is a variant of InfoNCE~\cite{CPC}, to maximize agreement between the two views. 
Besides, GCC~\cite{GCC} uses subgraph augmentation to define similar instances, and GCA~\cite{GCA} adopts adaptive graph augmentation to construct better graph views. 
Such learning paradigms can enable the model to effectively capture the common characteristics of data and improve model performance.

The current attempt of applying GCL on temporal graphs is usually to construct time-related graph views. 
Among them, DDGCL~\cite{DDGCL} uses timestamps to induce different subgraphs of a central node as two temporal graph views for contrastive learning, which enables TGNs to learn robust common features. 
Besides, time-dependent graph data augmentation (GDA) is also used to assist TGNs in learning more effective representations. 
For example, MeTA~\cite{MeTA} proposes to conduct graph data augmentation by adding existing duplicate edges while perturbing the timestamps on the edges, whose augmented graph structures are used for training. 
Unfortunately, the utilization of these methods for constructing temporal views or augmenting graphs is suboptimal as they lack learnability and fail to leverage rich information from downstream tasks.

\subsection{Graph Structure Learning}

Considering the flaws in the graph structure, graph structure learning (GSL) has been proposed to refine the original graph structure. 
Generally, GSL approaches use initial node features and an optional graph structure to construct the graph. A "structure learner" is employed to compute the edge weights for constructing the graph structure~\cite{CGI, SUBLIME, SLAPS, GRCN, ProGNN, IDGL, AD-GCL, AutoGCL}. Additional post-processing steps, such as discrete sampling, may be involved to produce the final graph. 
The refined graph structure is empirically better than the original one and can be used for further training and inference. 
Current GSL methods can be broadly classified into metric-based methods, neural methods, and direct methods~\cite{GSL_Survey}. 
Metric-based methods use non-learnable kernel functions like Gaussian or diffusion kernel to calculate reconstructed edge weights~\cite{GRCN, IDGL}. 
Neural methods utilize complex deep neural networks to compute edge weights, resulting in improved performance~\cite{CGI, SUBLIME, SLAPS, AD-GCL, AutoGCL}. 
Direct methods treat the adjacency matrix as free parameters, making them more flexible but challenging to optimize~\cite{ProGNN}.

Among these methods, AD-GCL~\cite{AD-GCL} parameterizes the edge-dropping augmentation by Gumble-Softmax~\cite{Gumble-softmax} and adopts a min-max principle for contrastive learning. 
CGI~\cite{CGI} employs Gumble-Softmax~\cite{Gumble-softmax} to parameterize the Bernoulli distribution and utilizes information bottleneck~\cite{Information_bottleneck} in contrastive learning to retain the necessary task-related information. 
Additionally, AutoGCL~\cite{AutoGCL} further adaptively chooses different structure perturbations by Gumble-Softmax and jointly trains with the model. 
Nevertheless, current GSL approaches are designed for static graphs and do not consider the important time-related information on temporal graphs.

\section{Preliminaries}
\label{sec:preliminaries}

\subsection{Notations and Problem Definition}
\label{sec:notations}

In this section, we give a brief description of the notations used in this paper. A temporal graph (here refers to CTDG) is denoted by $\mathcal{G}=\{\mathcal{V}, \mathcal{E}, \mathrm{X}\}$, where $\mathcal{V}$ is the node set, $\mathcal{E}$ is the edge set, $\mathrm{X} = (\mathrm{X}^{\mathcal{V}}, \mathrm{X}^{\mathcal{E}})$ is initial features containing node features $\mathrm{X}^{\mathcal{V}} \in \mathbb{R}^{|\mathcal{V}| \times D_{\mathcal{V}}}$ and edge features $\mathrm{X}^{\mathcal{E}} \in \mathbb{R}^{|\mathcal{E}| \times D_{\mathcal{E}}}$. The edge set can be denoted as $\mathcal{E} = \{(u_1, v_1, t_1, e_1), \cdots, (u_{|\mathcal{E}|}, v_{|\mathcal{E}|}, t_{|\mathcal{E}|}, e_{|\mathcal{E}|})\}$, each of which indicates the interaction happened between source node $u_i \in \mathcal{V}$ and destination node $v_i \in \mathcal{V}$ at timestamp $t_i$ with an associated feature $e_i \in \mathrm{X}^{\mathcal{E}}$. The one-hop neighbors of node $v$ are represented by $N(v)$, and we use $N^k(v)$ to represent the $k$-th hop neighbors. Though temporal graphs are multigraph, we use $uv$ to represent the index of edge(s) between node $u$ and $v$ for simplicity.

% problem definition
Our objective is to train a temporal graph network, denoted as $f$, which can encode the node's neighborhood information at a timestamp. 
This is accomplished by utilizing all the relevant interactions available on the temporal graph before this timestamp. The encoded embeddings of two arbitrary nodes are used to determine whether an interaction will occur between them at the given timestamp.

\subsection{Graph Neural Networks}
\label{sec:preliminaries_gnn}

Graph Neural Networks (GNNs)~\cite{GCN} show powerful representation ability for graph data. 
The node embeddings in GNNs are iteratively updated by aggregating the information from their neighboring nodes. 
Generally, the $l$-th layer of GNNs can be divided into two steps (i.e., message computation and aggregation):
\begin{equation}
\mathbf{h}_{v}^{(l)} = \mathrm{AGG}^{(l)}\left(\mathbf{h}_{v}^{(l-1)}, \mathrm{MSG}^{(l)}\left(\{\mathbf{h}_{u}^{(l-1)}, u \in N(v)\}; \theta_m^l \right); \theta_a^l \right)
\end{equation}
where $\mathbf{h}_{u}^{(l)} \in \mathbb{R}^{d_{l}}$ are the embedding vectors of nodes $u$ in layer $l$ and the embedding dimension is $d_l$. 
% $\mathbf{m}_{u}^{(l)}$ is the computed message for node $u$ in layer $l$. 
$\mathrm{MSG}^{(l)}(\cdot)$ is a message computation function parameterized by $\theta_m^l$ and 
$\mathrm{AGG}^{(l)}(\cdot)$ is a message aggregation function parameterized by $\theta_a^l$ in layer $l$.

\section{Time-aware Graph Structure Learning}
\label{sec:method_TGSL}

\subsection{Overview Framework}
\label{sec:method_overview}

\begin{figure*}[t]
	\centering
	\includegraphics[width=0.98\linewidth]{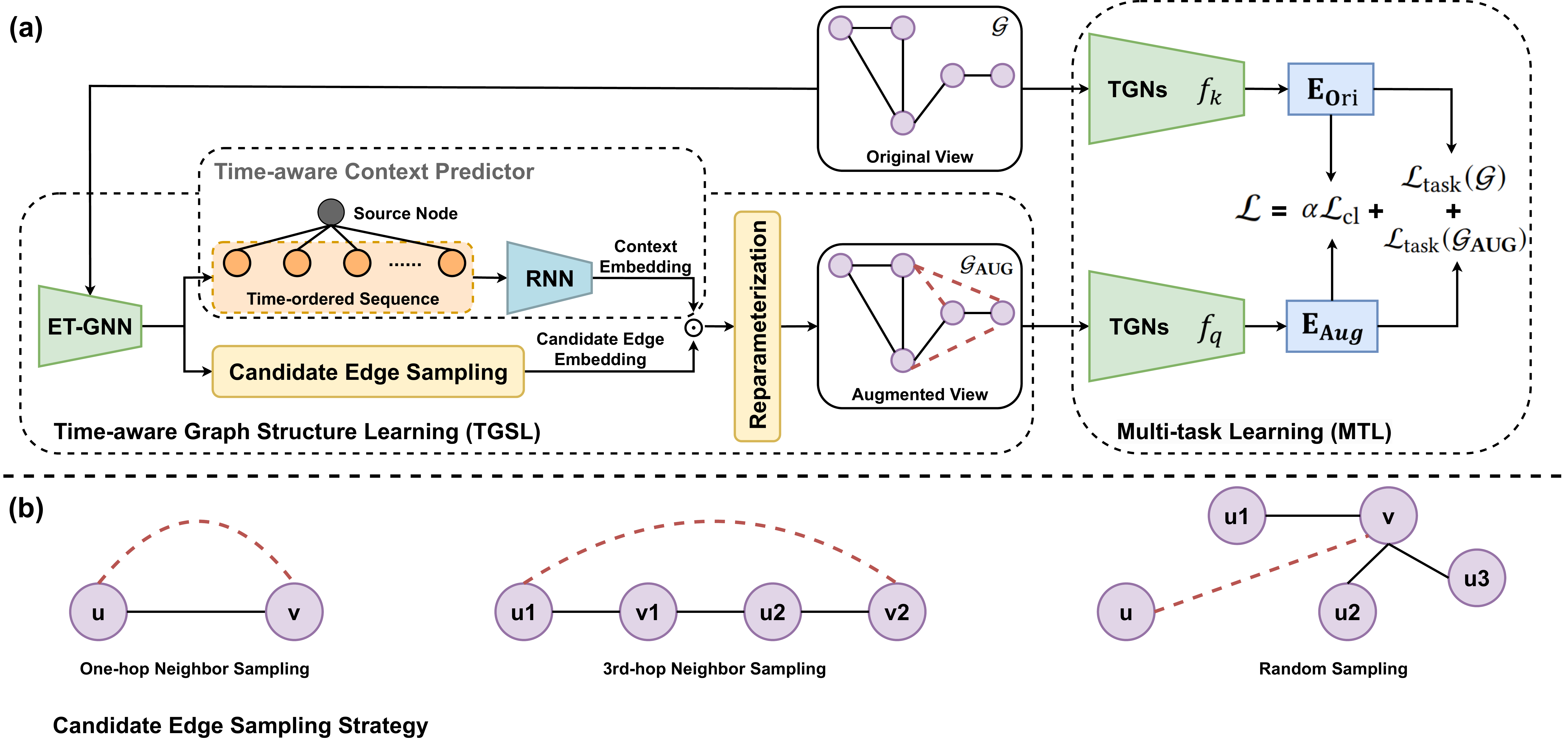}
% 	\vspace{-1ex}
	\caption{(a) TGSL Overview Framework. The original graph is first fed into ET-GNN extracting edge embeddings, which are further utilized by Time-aware Context Predictor and Candidate Edge Sampling to construct the augmented view. $\mathbf{E}_{\textbf{Ori}}$ and $\mathbf{E}_{\textbf{Aug}}$ represents the node embeddings output by TGNs $f_k$ and $f_q$, respectively. (b) Candidate Edge Sampling Strategy. }
% 	\vspace{-1ex}
	\label{model_figure}
\end{figure*}

Most existing GSL methods are designed for static graphs without considering temporal information on the graph. This motivates us to explore GSL on temporal graphs. The key aspect of conducting GSL on temporal graphs is understanding the fundamental difference between a temporal graph and a static graph. In a temporal graph, interactions are timestamped, meaning that the neighbors of nodes form a time-ordered sequence. This allows us to leverage RNNs to process interactions sequentially and generate time-aware context embeddings at a timestamp. Using these context embeddings, we can predict the nodes that are likely to interact at a given timestamp and add these selected interactions into the graph to refine the original flawed graph structure.

Figure \ref{model_figure} illustrates our overall framework, which consists of two main components: the Time-aware Graph Structure Learning (TGSL) module and the multi-task learning (MTL) module. The TGSL module is responsible for learning better graph structures by adding potential edges. It uses an edge-centric time-aware graph neural network (ET-GNN) for edge embedding extraction and uses an RNN to extract time-aware context embeddings. These context embeddings are then used to calculate edge weights for candidate edges which are sampled by diverse strategies, and the reparameterization trick Gumble-Top-K~\cite{Gumble-top-k} is employed to select the final K closest edges for augmentation. 
In the MTL module, two supervised losses on the original and learned graphs, as well as the contrastive loss between these two views, are used to optimize TGSL and TGNs jointly. This module aims to enhance the learning process by leveraging both the original and learned graph structure. 
The following subsections provide a detailed explanation of each of these crucial components.

\subsection{Edge-centric Time-aware Graph Neural Network (ET-GNN)}
\label{sec:method_embedding}

Since interactions are very critical and informative on temporal graphs, we argue that edge-centric message aggregation should be emphasized. By performing message aggregation on the edges, edge embeddings capture rich semantic information and can be optimized effectively. 
Moreover, considering the significance of temporal information on the edges of the temporal graph~\cite{TGAT}, we also integrate the time encoding~\cite{TE} in the message computation, which allows the GNN to adaptively learn the importance of different timestamps on the edges. 
As suggested by~\cite{GraphMixer}, because of the large numerical differences between different timestamps, we set the time encoding in TGSL to be non-learnable to ensure gradient stabilization during training. 
The time encoding used in ET-GNN can be formalized as a d-dimensional vector: 
\begin{equation}
\mathrm{TE}(t) = \cos \left(t \boldsymbol{\omega}\right)
\end{equation}
where $t$ is the timestamp and $\boldsymbol{\omega}=\left\{\alpha^{-(i-1) / \beta}\right\}_{i=1}^d$. $\alpha$ and $\beta$ are hyper-parameters and set to satisfy $\alpha=\beta=\sqrt{d}=10$ by default. 

Formally, the ET-GNN can be described by: 

\begin{equation}
\begin{aligned}
\mathbf{m}_{N(v)}^{(l)} &= \sum_{u \in N(v)} \frac{\operatorname{CONCAT}\left(\mathbf{h}_{u}^{(l-1)}, \mathbf{f}_{uv}^{(l-1)}, \mathrm{TE}(t_{uv})\right)}{|N(v)|} \\
\mathbf{h}_{v}^{(l)} &= \sigma\left(\mathbf{w}_h^{(l)} \cdot \operatorname{CONCAT}\left(\mathbf{h}_{v}^{(l-1)}, \mathbf{m}_{N(v)}^{(l)}\right)\right) \\
\mathbf{f}_{uv}^{(l)} &= \sigma\left(\mathbf{w}_f^{(l)} \cdot \operatorname{CONCAT}\left(\mathbf{f}_{uv}^{(l-1)}, \mathbf{h}_{u}^{(l-1)}, \mathbf{h}_{v}^{(l-1)}, \mathrm{TE}(t_{uv})\right)\right)
\end{aligned}
\end{equation}
where $\mathbf{f}_{uv}^{(l)} \in \mathbb{R}^{d_{l}}$ are the embedding vectors of the edge between node $u$ and $v$ in layer $l$ and $t_{uv}$ is the corresponding timestamp. 
$|N(v)|$ is the number of one-hop neighbors of node $v$, $\mathbf{w}_h^{(l)}, \mathbf{w}_f^{(l)} \in \mathbb{R}^{d_{l-1} \times d_l}$ is the parameter in the $l$-th layer, $\operatorname{CONCAT}(\cdot)$ denotes the concatenation operation and $\sigma(\cdot)$ denotes the ReLU activation function. 
Notably, the inputs of ET-GNN are given by initial node and edge feature vectors, i.e., $\mathbf{h}_{v}^{(0)}=\mathrm{X}_v^{\mathcal{V}}, \mathbf{f}_{uv}^{(0)}=\mathrm{X}_{uv}^{\mathcal{E}}$ and we directly use the output $\mathbf{f}_{uv}^{(L)}$ in final layer $L$ for further computation.

\subsection{Time-aware Context Predictor}
\label{sec:method_context}

In TGSL, we treat the nodes in each training batch as source nodes and sample destination nodes for them to form candidate edges. 
However, the question remains: how do we select the final edges to be added to the graph structure? 
Intuitively, we may first assign a weight to each candidate edge, which can be adaptively learned by TGSL. 
The edge weights are determined by the source and the sampled destination nodes, so we first extract time-aware context embedding for each source node, which represents the neighborhood information. 
Since each node's one-hop neighbors can be viewed as a time-ordered sequence, we use an RNN to predict the time-aware context embedding by feeding in the corresponding edge embeddings $\mathbf{f}_{uv}^{(L)}$ from ET-GNN. 
The time-aware context embedding well describes a node's state at a specific timestamp, and it can be helpful in selecting suitable candidate edges. 
We use the edge embedding that corresponds to the neighbor instead of the node embedding to remain compatible with ET-GNN and the candidate edge embeddings introduced later. 
Additionally, the edge embedding $\mathbf{f}_{uv}^{(L)}$ contains rich time-related information. 
The RNN can be any sequence model like long short-term memory (LSTM)~\cite{LSTM} or transformer~\cite{Self-Attention}, and we adopt LSTM in TGSL by default. 
The time-aware context embedding of node $u$ can be described as: 
\begin{equation}
\mathbf{z}_u = \operatorname{RNN}(\mathbf{f}_{uv_1}^{(L)}, \mathbf{f}_{uv_2}^{(L)}, \cdots, \mathbf{f}_{uv_{N_{\text{RNN}}}}^{(L)})
\end{equation}
where $N_{\text{RNN}}$ is the max input length of the RNN, $t_{uv_1} \leq t_{uv_2} \leq \cdots \leq t_{uv_{N_{\text{RNN}}}}$ and $v_1, v_2, \cdots, v_{N_{\text{RNN}}} \in N(u)$.

\subsection{Candidate Edge Construction}
\label{sec:method_edge_construction}

\subsubsection{Candidate Edge Sampling Strategy}
After obtaining the time-aware context embedding of source nodes, a reasonable destination node sampling strategy should be designed to construct the candidate edges. 
While many existing GSL methods use a pair-wise calculation approach to model the edge weights of all node pairs, this can be memory-intensive and inefficient. 
Instead, we argue that the key insight in designing the sampling strategies is that we should attempt to enrich the diversity of destination nodes while maintaining efficiency in the sampling process. 
Therefore, we propose the following three strategies for destination node sampling. 
\textbf{(1) One-hop neighbor sampling}. 
It is common for an entity to interact with its neighbors multiple times at different timestamps, and there may be some missing links between the entity and its neighbors, so we sample one-hop neighbors of source nodes and assign different timestamps to construct candidate edges. 
We use $\mathbf{f}_{uv}^{(L)}$ as the candidate edge features where $u$ and $v$ are the source and destination nodes. 
\textbf{(2) 3rd-hop neighbor sampling}. 
The one-hop neighbor sampling strategy constructs duplicate candidate edges from existing links. However, there are still some undetectable missing links between nodes that have never interacted with each other, and the one-hop neighbor sampling strategy does not help alleviate the situation. 
Instead, we can sample some neighbors of the nodes similar to source nodes. 
Here we directly sample the 3rd-hop neighbors of source nodes, as Figure \ref{model_figure} shows. 
If nodes $u_1$ and $u_2$ share common one-hop neighbors like $v_1$, they can be considered similar to some extent, and we can use the node $u_2$'s one-hop neighbor $v_2$ as the destination node. 
Especially, we use $\mathbf{f}_{u_2v_2}^{(L)}$ as candidate edge features, which are "borrowed" from edge $u_2v_2$. 
\textbf{(3) Random sampling}. 
Although the first two sampling strategies alleviate the missing link problem, we cannot efficiently reach the long-distance missing links by sampling the $n$-th-hop neighbors due to the sampling costs. 
Therefore, we propose randomly sampling destination nodes from the training set to further enhance candidate edges' diversity. 
In random sampling, the candidate edge features are initialized as zero embedding vectors. 
Notably, we sample at most $N_{\text{CAN}}$ destination nodes for each source node.

\subsubsection{Time-mapping Mechanism}

So far, we have obtained the source nodes, destination nodes, and features of candidate edges. 
On temporal graphs, there must have a timestamp on each edge. For simplicity, we just randomly sample a timestamp $t_{\text{new}}$ from the range 0 to $t_{\text{max}}$, which is the maximum timestamp in the training data. 
Early works~\cite{JODIE} point out that the node's embedding will drift farther in its embedding space as more time elapse. 
However, the corresponding timestamps of the context embedding and candidate edge feature, which is $t_{\text{max}}$ and the sampled neighbor's timestamp $t_{\text{sample}}$, are different from the newly sampled timestamp. 
Inspired by this, we propose to project the context embedding and the candidate edge embedding to the newly sampled timestamp through a time-mapping mechanism. 
To achieve this, we input the difference value between the old timestamp and the newly sampled timestamp, denoted as $\Delta$, into a time-difference encoder, which generates a time-context vector $\mathbf{s}(\Delta)$ for both the context embedding and candidate edge feature. 
This can be expressed as follows: 
\begin{equation}
\mathbf{s}(\Delta) = \sin \left(\Delta \boldsymbol{\omega}\right) + \mathbf{1}
\end{equation}
where $\boldsymbol{\omega}$ is a non-learnable parameter introduced in Section \ref{sec:method_embedding}. When $\Delta = 0$, $\mathbf{s}(\Delta) = \mathbf{1}$, which represents the identity mapping. 
The context embedding and candidate edge feature will be projected to the newly sampled timestamp by element-wise product with the corresponding time-context vector: 
\begin{equation}
\mathbf{\hat{z}}_u = \mathbf{z}_u \otimes \mathbf{s}(t_{\text{new}} - t_{\text{max}})
\end{equation}
\begin{equation}
\mathbf{\hat{f}}_{uv}^{(L)} = \mathbf{f}_{uv}^{(L)} \otimes \mathbf{s}(t_{\text{new}} - t_{\text{sample}})
\end{equation}
where $\otimes$ denotes element-wise product, $\mathbf{\hat{z}}_u$ and $\mathbf{\hat{f}}_{uv}^{(L)}$ are projected embeddings. 
Notably, we do not use the absolute difference value because the embedding can be projected to a past timestamp as well as a future timestamp, and the sign of the difference value can just implicitly distinguish the two cases. 
With the time-mapping mechanism, the embedding can be compatible with any sampled timestamp and evolve over time.

\subsubsection{Candidate Edge Selection Based on Reparameterization (i.e., Gumble-Top-K)}

After the time-mapping operations, we obtain the projected context embedding $\mathbf{\hat{z}}_u$ and candidate edge features $\mathbf{\hat{f}}_{uv}^{(L)}$. 
Adding all the candidate edges to the original graph will greatly increase the computational overhead of TGNs, and due to the randomness of the candidate edge sampling, not every candidate edge should be added. 
Intuitively, we may assign a learnable weight to each candidate edge, which can be adaptively optimized by the model. 
A naive solution is to select $K$ candidate edges with the largest weights, but it will hinder the efficient exploration of other candidate edges because the selection process here is a deterministic process. 
Therefore, we parameterize the selection process of candidate edges as a learnable sampling process. 
Each candidate edge is associated with a discrete random variable $\mathbf{\rho}_{uv} \sim \text{Bernoulli}(\mathbf{m}_{uv})$, indicating whether to select the edge. 
$\mathbf{m}_{uv}$ derives from the dot product of the context embedding and candidate edge feature: 
\begin{equation}
\mathbf{m}_{uv} = \mathbf{\hat{z}}_u^T \odot 
\mathbf{\hat{f}}_{uv}^{(L)}
\end{equation}

To optimize TGSL in an end-to-end manner, we adopt the Gumble-Top-K trick ~\cite{Gumble-top-k} and relax $\mathbf{\rho}_{uv}$ to be a continuous variable in $[0, 1]$: 
\begin{equation}
\mathbf{\rho}_{uv}=\text{Sigmoid}((\log u-\log (1-u)+\mathbf{m}_{uv}) / \tau)
\end{equation}
where $u \sim \text{Uniform}(0, 1)$ and $\tau \in \mathbb{R}^{+}$ denotes temperature coefficient. The gradient $\frac{\partial \mathbf{\rho}_{uv}}{\partial \mathbf{m}_{uv}}$ is well-defined, and the sampling process can be directly optimized through backpropagation smoothly, making training more efficient. 
For a source node $u$ and its candidate destination node $v_i \in \{v_1, v_2, \cdots, v_{\text{CAN}}\}$, we select $K$ candidate edges with the largest $\mathbf{\rho}_{uv_i}$. 
It is worth noting that $\mathbf{\rho}_{uv}$ also serves as the edge weight in TGNs to ensure gradient backpropagation. 
By doing so, the edges for final addition can be explored diversely in the candidate edge set, resulting in enhanced model performance.

\subsection{End-to-End Training and Inference with TGNs}
\label{sec:method_training}
We jointly optimize TGNs and TGSL in an end-to-end manner through the multi-task learning (MTL) of three objectives. 
We adopt graph contrastive learning using two different graph views: the original graph, denoted as $\mathcal{G}$, and the edge-addition augmented graph, denoted as $\mathcal{G}_{\textbf{AUG}}$.
The original graph acts as an anchor view for stable training, and TGNs serve as the encoder denoted by $f$, as Figure \ref{model_figure} shows. 
Especially, we adopt the MoCo training strategy~\cite{MoCo} instead of the End-to-End (E2E) contrastive strategy to make full use of the negative samples. 
InfoNCE~\cite{CPC} is utilized to calculate the loss of contrastive learning, which can be described as: 
\begin{equation}
\mathcal{L}_{\mathrm{cl}} = -\log \frac{\exp \left(\boldsymbol{q}^{\top} \boldsymbol{k}_{+} / \tau\right)}{\sum_{i=0}^M \exp \left(\boldsymbol{q}^{\top} \boldsymbol{k}_i / \tau\right)}
\end{equation}
where $M$ is the number of negative samples and $\tau$ is the temperature coefficient. 
$\boldsymbol{q} = f_q(\mathcal{G}_{\textbf{AUG}})$ is an encoded query vector output from the query encoder $f_q$ and $\boldsymbol{k} = f_k(\mathcal{G})$ is an encoded key vector output from the key encoder $f_k$. 
Notably, $\boldsymbol{k}_{+}$ denotes the single positive key for the query $\boldsymbol{q}$. 

For the training scheme, we adopt MTL to jointly optimize the supervised task (i.e., temporal link prediction) with the binary cross entropy loss function denoted as $\mathcal{L}_{\mathrm{task}}$, and the self-supervised contrastive learning task with loss function $\mathcal{L}_{\mathrm{cl}}$. 
We also conduct the supervised task on the augmented graph $\mathcal{G}_{\textbf{AUG}}$ for better structure refinement and representation learning. 
The total loss can be formally written as the following: 
\begin{equation}
\mathcal{L} = \mathcal{L}_{\mathrm{task}}(\mathcal{G}) + \mathcal{L}_{\mathrm{task}}(\mathcal{G}_{\textbf{AUG}}) + \alpha\mathcal{L}_{\mathrm{cl}}
\end{equation}
where $\alpha \in [0, 1]$ is a balance hyper-parameter. 

During the evaluation, we directly use the well-trained TGNs and the augmented graph $\mathcal{G}_{\textbf{AUG}}$ for inference.

\section{Experiments}
\label{sec:exp}

\subsection{Experimental Setup}

\begin{table*}[t]
  % \footnotesize
  \caption{Test ACC and AP of TRANSDUCTIVE Temporal Link Prediction. The results are reported as the mean (\%) $\pm$ standard deviation over 3 runs with different seeds. We bold the superior results on TGAT and GraphMixer, respectively, and also bold the best results among all methods if they occur with other TGNs. $^*$ denotes the non-learnable time encoding is applied. }
%   \vspace{-1ex}
  \label{tab:transductive}
%   \resizebox{\linewidth}{!}{
  \begin{tabular}{c|cc|cc|cc}
    \toprule
    Dataset & \multicolumn{2}{c|}{Wikipedia} & \multicolumn{2}{c|}{Escorts} & \multicolumn{2}{c}{Reddit} \\
    \midrule
    Model & ACC & AP & ACC & AP & ACC & AP \\
    \midrule
    % Results
    JODIE~\cite{JODIE}
    & 88.36 $\pm$ 0.6 
    & 95.59 $\pm$ 0.4 
    & 80.52 $\pm$ 0.7 
    & 89.31 $\pm$ 0.6
    & 92.36 $\pm$ 0.4 
    & 97.78 $\pm$ 0.1 \\
    DyRep~\cite{DyRep}
    & 86.26 $\pm$ 0.3 
    & 94.34 $\pm$ 0.1 
    & 77.25 $\pm$ 0.3 
    & 85.64 $\pm$ 0.3
    & 92.52 $\pm$ 0.1 
    & 97.91 $\pm$ 0.0 \\
    TGN~\cite{TGN}
    & 92.25 $\pm$ 0.2 
    & 98.07 $\pm$ 0.1 
    & 78.18 $\pm$ 0.6 
    & 87.58 $\pm$ 0.4
    & 93.75 $\pm$ 0.2 
    & 98.48 $\pm$ 0.1 \\
    % \midrule
    TGAT~\cite{TGAT}
    & 87.97 $\pm$ 0.5 
    & 95.50 $\pm$ 0.3 
    & 72.49 $\pm$ 0.5 
    & 80.71 $\pm$ 0.3
    & 92.89 $\pm$ 0.2 
    & 98.15 $\pm$ 0.1 \\
    \midrule
    TGAT*~\cite{TGAT}
    & 90.31 $\pm$ 0.3 
    & 96.90 $\pm$ 0.2 
    & 79.85 $\pm$ 0.2 
    & 87.54 $\pm$ 0.3
    & 93.06 $\pm$ 0.0 
    & 98.24 $\pm$ 0.0 \\
    \textbf{TGAT* + TGSL}
    & \textbf{93.29 $\pm$ 0.1} 
    & \textbf{98.19 $\pm$ 0.1} 
    & \textbf{80.68 $\pm$ 0.2} 
    & \textbf{88.01 $\pm$ 0.3}
    & \textbf{94.31 $\pm$ 0.1} 
    & \textbf{98.67 $\pm$ 0.0} \\
    \midrule
    GraphMixer~\cite{GraphMixer}
    & 89.76 $\pm$ 0.2 
    & 96.65 $\pm$ 0.2 
    & 82.57 $\pm$ 0.1 
    & 91.19 $\pm$ 0.1
    & 90.27 $\pm$ 0.2 
    & 96.69 $\pm$ 0.1 \\
    \textbf{GraphMixer + TGSL}
    & \textbf{90.83 $\pm$ 0.5} 
    & \textbf{97.19 $\pm$ 0.4} 
    & \textbf{84.05 $\pm$ 1.5} 
    & \textbf{91.60 $\pm$ 0.7} 
    & \textbf{90.68 $\pm$ 0.0} 
    & \textbf{96.97 $\pm$ 0.0} \\
    \bottomrule
  \end{tabular}
%   }
\end{table*}

\begin{table*}[t]
  % \footnotesize
  \caption{Test ACC and AP of INDUCTIVE Temporal Link Prediction. The results are reported as the mean (\%) $\pm$ standard deviation over 3 runs with different seeds. We bold the superior results on TGAT and GraphMixer respectively, and also bold the best results among all methods if they occur with other TGNs. $^*$ denotes the non-learnable time encoding is applied.}
%   \vspace{-1ex}
  \label{tab:inductive}
%   \resizebox{\linewidth}{!}{
  \begin{tabular}{c|cc|cc|cc}
    \toprule
    Dataset & \multicolumn{2}{c|}{Wikipedia} & \multicolumn{2}{c|}{Escorts} & \multicolumn{2}{c}{Reddit} \\
    \midrule
    Model & ACC & AP & ACC & AP & ACC & AP \\
    \midrule
    % Results
    JODIE~\cite{JODIE}
    & 84.54 $\pm$ 0.9 
    & 93.35 $\pm$ 0.5 
    & 64.94 $\pm$ 0.6 
    & 75.40 $\pm$ 0.7
    & 88.94 $\pm$ 0.6 
    & 95.50 $\pm$ 0.3 \\
    DyRep~\cite{DyRep}
    & 82.32 $\pm$ 0.1 
    & 91.67 $\pm$ 0.1 
    & 64.16 $\pm$ 0.4 
    & 69.91 $\pm$ 0.1
    & 88.60 $\pm$ 0.5 
    & 95.31 $\pm$ 0.3 \\
    TGN~\cite{TGN}
    & \textbf{90.53 $\pm$ 0.1} 
    & \textbf{97.42 $\pm$ 0.1} 
    & 64.47 $\pm$ 1.9 
    & 74.26 $\pm$ 1.0
    & 90.55 $\pm$ 0.5 
    & 96.97 $\pm$ 0.1 \\
    % \midrule
    TGAT~\cite{TGAT}
    & 85.47 $\pm$ 0.1 
    & 93.95 $\pm$ 0.1 
    & 63.68 $\pm$ 0.3 
    & 69.38 $\pm$ 0.0
    & 91.39 $\pm$ 1.2 
    & 97.40 $\pm$ 0.6 \\
    \midrule
    TGAT*~\cite{TGAT}
    & 88.05 $\pm$ 0.4 
    & 95.66 $\pm$ 0.2 
    & 65.54 $\pm$ 0.4 
    & 74.21 $\pm$ 0.3
    & 91.51 $\pm$ 1.0 
    & 97.42 $\pm$ 0.6 \\
    \textbf{TGAT* + TGSL}
    & \textbf{89.90 $\pm$ 0.3} 
    & \textbf{96.70 $\pm$ 0.1} 
    & \textbf{67.55 $\pm$ 1.1} 
    & \textbf{75.40 $\pm$ 0.2}
    & \textbf{91.77 $\pm$ 1.4} 
    & \textbf{97.50 $\pm$ 0.6} \\
    \midrule
    GraphMixer~\cite{GraphMixer}
    & 88.25 $\pm$ 0.3 
    & 96.12 $\pm$ 0.2 
    & 63.92 $\pm$ 1.8 
    & 77.01 $\pm$ 0.1
    & 87.24 $\pm$ 0.2 
    & 94.67 $\pm$ 0.2 \\
    \textbf{GraphMixer + TGSL}
    & \textbf{89.64 $\pm$ 0.3} 
    & \textbf{96.56 $\pm$ 0.4} 
    & \textbf{65.64 $\pm$ 0.6} 
    & \textbf{77.51 $\pm$ 0.6}
    & \textbf{87.73 $\pm$ 0.0} 
    & \textbf{95.06 $\pm$ 0.0} \\
    \bottomrule
  \end{tabular}
%   }
\end{table*}

\subsubsection{Datasets and Evaluation Tasks}

We adopt three datasets for evaluation: Wikipedia~\cite{Datasets} is a bipartite graph containing the interaction of users editing pages. Each edge in the graph is linked to a 172-dimensional embedding vector. 
Reddit~\cite{Datasets} is a bipartite graph describing users' post behavior on subreddits, and a 172-dimensional embedding vector is also given on each edge. 
Escorts~\cite{network_repo} is a bipartite network of sex buyers and their escorts. 
Following early works~\cite{TGAT, TGN}, we adopt the chronological split with a train-validation-test ratio of 70\%-15\%-15\%. 

We compare our proposed method with several competitive baselines on the temporal link prediction task under both transductive and inductive settings for a comprehensive evaluation. The edges used for evaluation under the transductive setting are among the nodes observed during training, while under the inductive setting, the evaluation edges are among unseen nodes ~\cite{TGAT}. For evaluation metrics, we report ACC (accuracy) and AP (average precision) following~\cite{MeTA}.

\subsubsection{Implementation Details}
\label{exp:setup}

Because our proposed method is based on temporal graphs, we adopt TGAT~\cite{TGAT} and GraphMixer~\cite{GraphMixer} to serve as the encoder of our proposed TGSL. 
We follow the official implementation of TGAT~\cite{TGAT}. 
As for GraphMixer~\cite{GraphMixer}, we adopt the implementation of~\cite{DyGLib}, which removes the one-hot encoding of nodes to enable the inductive setting. 
As suggested in~\cite{GraphMixer}, the time encoding is non-learnable to ensure training stabilization, so we also apply this modification to TGAT~\cite{TGAT} for better performance. 
We set the batch size to 200 and adopt the Adam optimizer with a learning rate of 1e-4. 
We adopt early stopping in our implementation. Specifically, we train TGNs with a max epoch number of 50, and the patience and the tolerance are set to 3 and 1e-3 following~\cite{TGAT}. 
The rest of the hyper-parameters w.r.t. TGNs are set according to original papers~\cite{TGAT, GraphMixer}. 
As for our proposed TGSL, we select the best candidate edge sampling strategy for each dataset, and the number of the final selected edges $K$ is set to 8. 
The balance hyper-parameter $\alpha$ in total loss function $\mathcal{L}$ and the temperature coefficient $\tau$ in InfoNCE~\cite{CPC} is tuned over the range $[0.1, 1.0]$. 
The temperature coefficient $\tau$ in Gumble-Top-K~\cite{Gumble-top-k} is fixed to $1.0$, and the queue size used in MoCo contrastive learning strategy~\cite{MoCo} is set to 512. 
We implement TGSL with PyTorch and run each experiment 3 times with different seeds on an NVIDIA V100 GPU. 

\subsubsection{Baselines}
For comparison experiments, we adopt the following baselines: 
\begin{itemize}
\item \textbf{Temporal Graph Networks.} In addition to TGAT~\cite{TGAT} and GraphMixer~\cite{GraphMixer}, we also adopt other three temporal graph network baselines: JODIE~\cite{JODIE}, DyRep~\cite{DyRep}, and TGN~\cite{TGN}. The training configurations are consistent with the original paper. 
\item \textbf{Contrastive Learning-based Methods.} Except for the TGNs, there are several approaches using contrastive learning or graph augmentation to improve model performance. Hence, we also adopt the following baselines for comprehensive evaluation: GraphCL~\cite{GraphCL}, DDGCL~\cite{DDGCL}, MeTA~\cite{MeTA}, and AD-GCL~\cite{AD-GCL}, which can all be integrated with TGNs to improve performance like ours. Among them, AD-GCL is a GSL method that introduces a trainable edge-dropping graph augmentation. Due to its similarity to our approach, we consider it as our baseline method. 
\end{itemize}

To ensure a fair comparison, we all employ end-to-end training with the same multi-task learning losses for all contrastive learning methods, focusing on the comparison of the augmented graph views $\mathcal{G}_{\textbf{AUG}}$. 
For GraphCL~\cite{GraphCL}, we use edge addition instead of edge perturbation, which corresponds to our proposed method. For MeTA~\cite{MeTA}, we implement the edge addition with time perturbation as the graph view. For DDGCL~\cite{DDGCL}, we directly use the introduced temporal view for contrastive learning. For AD-GCL~\cite{AD-GCL}, we use the edge-dropping view instead of edge addition in TGSL. 
Besides, we use the same number $K$ of edge addition for GraphCL~\cite{GraphCL} and MeTA~\cite{MeTA} and tune the other hyper-parameters carefully.

\subsection{Main Results}

\subsubsection{Compare to Temporal Graph Network Baselines}
The comparison results of temporal graph network baselines on the Wikipedia, Reddit, and Escorts datasets w.r.t. temporal link prediction under both transductive and inductive settings are shown in Table \ref{tab:transductive} and \ref{tab:inductive}. 
According to Table \ref{tab:transductive} and \ref{tab:inductive}, we can draw several conclusions. 
(1) TGSL improves both TGAT and GraphMixer by a considerable margin across all datasets under both transductive and inductive settings, which demonstrates the effectiveness of our proposed TGSL. 
(2) By equipping TGSL, TGAT and GraphMixer achieve the best results w.r.t. ACC and AP on the three datasets compared to the temporal graph network baselines under the transductive setting, which benefits from the superiority of the learned graph structure. 
(3) We observe slightly smaller improvements in the results on the Reddit dataset compared to other datasets. We think it could be due to the difference in node degree statistics displayed in Table \ref{tab:deg}. 
TGSL improves graph structure by adding edges, which is particularly beneficial for relatively sparse datasets, while Reddit's relatively large node degree may limit its impact. 
Further analysis on this will be provided in Section \ref{sec:missing_analysis}.

\begin{table}[H]
  % \footnotesize
% \vspace{-1ex}
  \caption{Node Degree Statistics of Training Datasets. }
%   \vspace{-1ex}
%   \small
  \label{tab:deg}
  \begin{tabular}{c|cccccc}
    \toprule
    Dataset & Min & Max & Mean & Q1 & Median & Q3 \\
    \midrule
    % Results
    Wikipedia
    & 1 & 1219 & 26 & 1 & 3 & 17 \\
    Reddit
    & 1 & 28904 & 76 & 21 & 30 & 49 \\
    Escorts
    & 1 & 572 & 5 & 1 & 2 & 4 \\
    \bottomrule
  \end{tabular}
%   \vspace{-1ex}
\end{table}

\begin{table*}[t]
  % \footnotesize
% \vspace{-1ex}
  \caption{Test ACC and AP of TGSL and Other Contrastive Learning-based Methods for Temporal Link Prediction on the Wikipedia Dataset. The results are reported as the mean (\%) $\pm$ standard deviation over 3 runs with different seeds. We bold the best results on TGAT and GraphMixer, respectively. $^*$ denotes the non-learnable time encoding is applied.}
%   \vspace{-1ex}
%   \small
  \label{tab:cl_baselines}
  \begin{tabular}{c|cc|cc}
    \toprule
    Setting & \multicolumn{2}{c|}{Transductive} & \multicolumn{2}{c}{Inductive} \\
    \midrule
    Model & ACC & AP & ACC & AP \\
    \midrule
    % Results
    TGAT*~\cite{TGAT}
    & 90.31 $\pm$ 0.3 
    & 96.90 $\pm$ 0.2 
    & 88.05 $\pm$ 0.4 
    & 95.66 $\pm$ 0.2 \\
    TGAT* + GraphCL (Add Edge)~\cite{GraphCL}
    & 90.88 $\pm$ 0.3 
    & 97.27 $\pm$ 0.0 
    & 88.59 $\pm$ 0.4 
    & 96.20 $\pm$ 0.2 \\
    TGAT* + MeTA (Add Edge \& Perturb Time)~\cite{MeTA}
    & 91.10 $\pm$ 0.1 
    & 97.37 $\pm$ 0.1 
    & 88.75 $\pm$ 0.4 
    & 96.32 $\pm$ 0.3 \\
    TGAT* + DDGCL (Temporal View)~\cite{DDGCL}
    & 91.16 $\pm$ 0.2 
    & 97.49 $\pm$ 0.1 
    & 88.88 $\pm$ 0.5 
    & 96.27 $\pm$ 0.2 \\
    TGAT* + AD-GCL (Drop Edge)~\cite{AD-GCL}
    & 90.79 $\pm$ 0.1 
    & 97.25 $\pm$ 0.0 
    & 88.91 $\pm$ 0.5 
    & 96.27 $\pm$ 0.3 \\
    \textbf{TGAT* + TGSL}
    & \textbf{93.29 $\pm$ 0.1} 
    & \textbf{98.19 $\pm$ 0.1} 
    & \textbf{89.90 $\pm$ 0.3} 
    & \textbf{96.70 $\pm$ 0.1} \\
    \midrule
    GraphMixer~\cite{GraphMixer}
    & 89.76 $\pm$ 0.2 
    & 96.65 $\pm$ 0.2 
    & 88.25 $\pm$ 0.3 
    & 96.12 $\pm$ 0.2 \\
    GraphMixer + GraphCL (Add Edge)~\cite{GraphCL}
    & 90.19 $\pm$ 0.1 
    & 96.83 $\pm$ 0.0 
    & 89.44 $\pm$ 0.2 
    & 96.39 $\pm$ 0.0 \\
    GraphMixer + MeTA (Add Edge \& Perturb Time)~\cite{MeTA}
    & 90.35 $\pm$ 0.2 
    & 96.92 $\pm$ 0.2 
    & 89.31 $\pm$ 0.1 
    & 96.39 $\pm$ 0.1 \\
    GraphMixer + DDGCL (Temporal View)~\cite{DDGCL}
    & 90.26 $\pm$ 0.0 
    & 96.95 $\pm$ 0.0 
    & 88.74 $\pm$ 0.1 
    & 96.26 $\pm$ 0.1 \\
    GraphMixer + AD-GCL (Drop Edge)~\cite{AD-GCL}
    & 90.40 $\pm$ 0.1 
    & 96.97 $\pm$ 0.1 
    & 89.35 $\pm$ 0.1 
    & 96.45 $\pm$ 0.1 \\
    \textbf{GraphMixer + TGSL}
    & \textbf{90.83 $\pm$ 0.5} 
    & \textbf{97.19 $\pm$ 0.4} 
    & \textbf{89.64 $\pm$ 0.3} 
    & \textbf{96.56 $\pm$ 0.4} \\
    \bottomrule
  \end{tabular}
%   \vspace{-1ex}
\end{table*}

\subsubsection{Compare to Contrastive Learning-based Baselines}
The comparison results of contrastive learning-based baselines on the Wikipedia dataset regarding temporal link prediction under both transductive and inductive settings are shown in Table \ref{tab:cl_baselines}. 
From Table \ref{tab:cl_baselines}, compared with the four baselines, i.e., GraphCL~\cite{GraphCL}, MeTA~\cite{MeTA}, DDGCL~\cite{DDGCL}, and AD-GCL~\cite{AD-GCL}, TGSL outperforms them in every aspect. 
GraphCL~\cite{GraphCL} employs random perturbations as graph augmentations to construct graph views, but the time-related information on temporal graphs is ignored, causing performance degradation. 
Although DDGCL~\cite{DDGCL} and MeTA~\cite{MeTA} take the time-related factors into account, the former essentially just removes some most recent edges to construct the temporal view, while the latter only considers uniformly adding repeated interactions, thereby limiting the graph representation learning. 
As for AD-GCL~\cite{AD-GCL}, contrary to TGSL, it uses edge dropping to construct graph views. However, edge dropping is only applied to existing edges and doesn't incorporate time-related information, limiting its potential to improve performance, so the effectiveness of edge dropping on temporal graphs is not as good as edge addition like ours.

\subsection{Ablation Studies}

In this section, we conduct ablation studies of TGSL on the Wikipedia dataset, and the results are shown in Table \ref{tab:ablation}. 
To verify the superiority of the learned graph structure, we compare it with the original graph structure for inference ('OGI') while keeping other settings unchanged.  
Besides, we alter the total loss function by removing the term $\mathcal{L}_{\mathrm{task}}(\mathcal{G}_{\textbf{AUG}})$ to verify the effectiveness and necessity of conducting supervised learning on the augmented graph. 
Moreover, we remove the ET-GNN and instead utilize the initial edge features of datasets followed by a multi-layer perception (MLP) to show the impact of the learned edge representation of ET-GNN.

From the results presented in Table \ref{tab:ablation}, several conclusions can be drawn. 
(1) The learned graph structure is obviously better than the original one. 
By using the learned graph structure for inference, the ACC and AP are increased by 2.55\% and 0.84\%, respectively, under the transductive setting and 1.07\% and 0.31\% under the inductive setting on the Wikipedia dataset. 
(2) Conducting supervised learning on the augmented graph contributes a lot to the model performance. 
Under the transductive setting, the ACC and AP are increased by 2.56\% and 0.90\% after integrating the supervised loss function on the augmented graph. Moreover, an obvious improvement can also be observed in the results under the inductive setting. 
(3) The edge representations learned by ET-GNN are more powerful than the initial edge features provided by the datasets, and the results are obviously improved after using ET-GNN. 
It is noteworthy that the results of all the above ablation experiments still surpass a single TGAT with a non-learnable time encoding module, which shows the robustness of TGSL.

\begin{table}[h]
  % \footnotesize
% \vspace{-1ex}
  \caption{Ablation Studies of TGSL on the Wikipedia Dataset. }
%   \vspace{-1ex}
%   \small
  \label{tab:ablation}
  \begin{tabular}{c|cc}
    \toprule
    Method \& Variants & ACC & AP \\
    \midrule
    \multicolumn{3}{c}{\textbf{Transductive}} \\
    \midrule
    TGAT*~\cite{TGAT}
    & 90.31 $\pm$ 0.3 
    & 96.90 $\pm$ 0.2 \\
    \midrule
    TGAT* + TGSL w/ OGI
    & 90.97 $\pm$ 0.3 
    & 97.37 $\pm$ 0.2 \\
    TGAT* + TGSL w/o $\mathcal{L}_{\mathrm{task}}(\mathcal{G}_{\textbf{AUG}})$
    & 90.96 $\pm$ 0.3 
    & 97.31 $\pm$ 0.1 \\
    TGAT* + TGSL w/o ET-GNN
    & 92.16 $\pm$ 0.2 
    & 97.73 $\pm$ 0.1 \\
    \midrule
    \textbf{TGAT* + TGSL}
    & \textbf{93.29 $\pm$ 0.1} 
    & \textbf{98.19 $\pm$ 0.1} \\
    \midrule
    \multicolumn{3}{c}{\textbf{Inductive}} \\
    \midrule
    TGAT*~\cite{TGAT}
    & 88.05 $\pm$ 0.4 
    & 95.66 $\pm$ 0.2 \\
    \midrule
    TGAT* + TGSL w/ OGI
    & 88.95 $\pm$ 0.2 
    & 96.40 $\pm$ 0.1 \\
    TGAT* + TGSL w/o $\mathcal{L}_{\mathrm{task}}(\mathcal{G}_{\textbf{AUG}})$
    & 89.00 $\pm$ 0.4 
    & 96.43 $\pm$ 0.2 \\
    TGAT* + TGSL w/o ET-GNN
    & 89.37 $\pm$ 0.4 
    & 96.47 $\pm$ 0.3 \\
    \midrule
    \textbf{TGAT* + TGSL}
    & \textbf{89.90 $\pm$ 0.3} 
    & \textbf{96.70 $\pm$ 0.1} \\
    \bottomrule
  \end{tabular}
%   \vspace{-1ex}
\end{table}

\subsection{More Analysis on the Learned Graph Structure}
\label{sec:analysis}

\subsubsection{Performance on Graphs with Different Sparsity}
\label{sec:missing_analysis}

\begin{figure*}[t]
% \vspace{-1ex}
    \centering
	\begin{subfigure}{0.24\linewidth}
		\centering
		\includegraphics[width=1.0\linewidth]{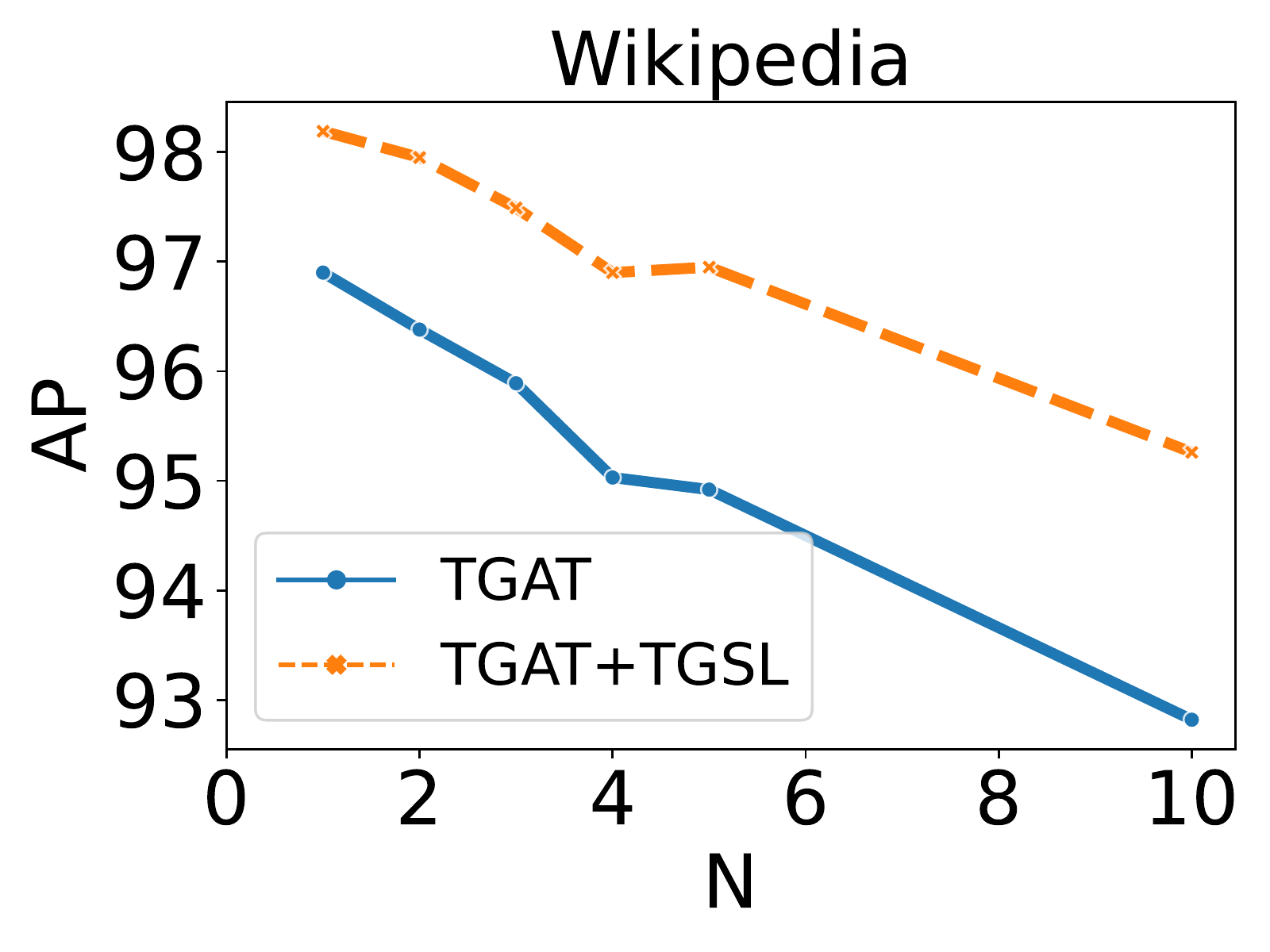}
		\caption{Transductive AP}
		\label{wiki_ap_trans}
	\end{subfigure}
	\centering
	\begin{subfigure}{0.24\linewidth}
		\centering
		\includegraphics[width=1.0\linewidth]{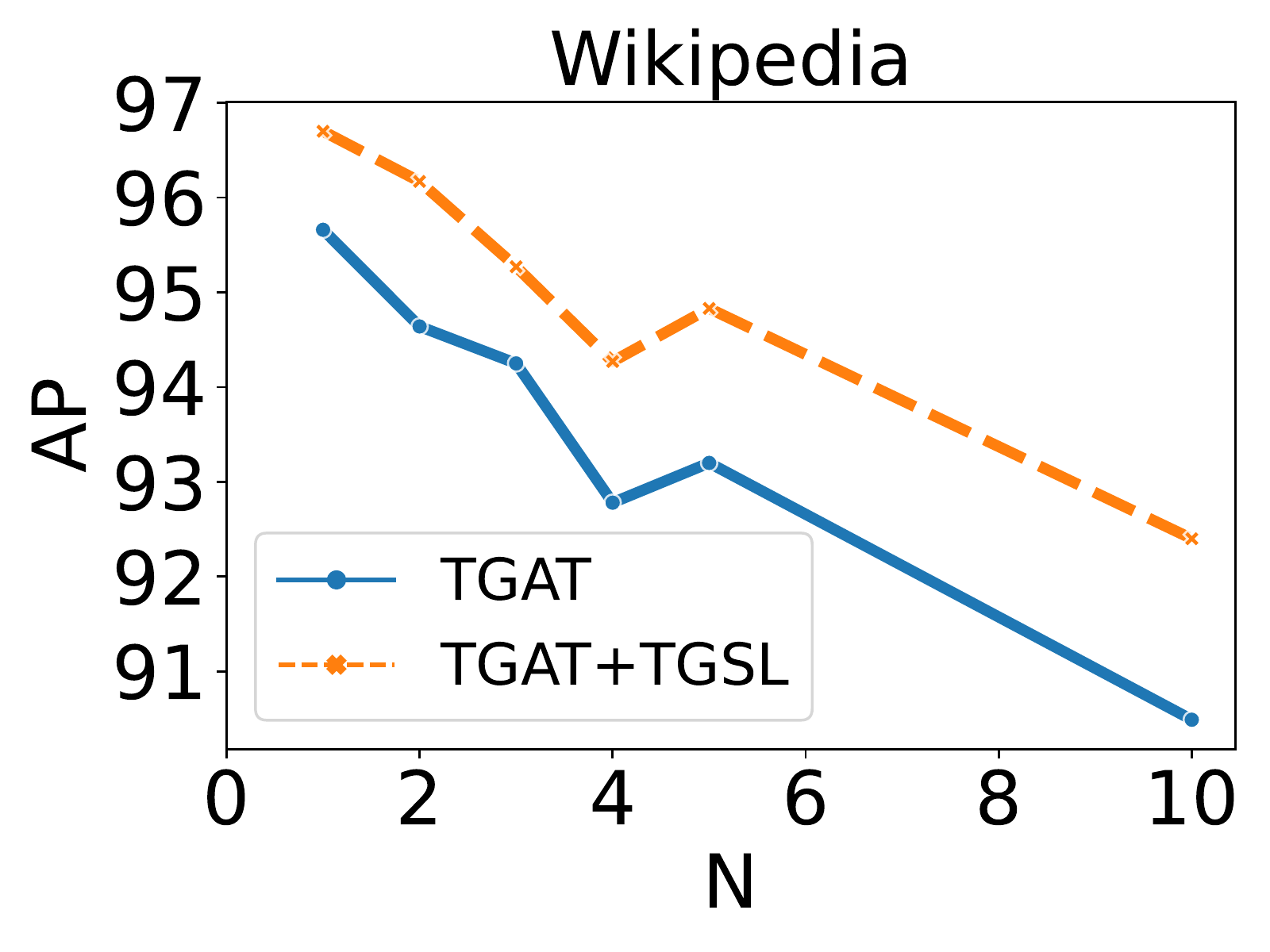}
		\caption{Inductive AP}
		\label{wiki_ap_induc}
	\end{subfigure}
	\centering
	\begin{subfigure}{0.24\linewidth}
		\centering
		\includegraphics[width=1.0\linewidth]{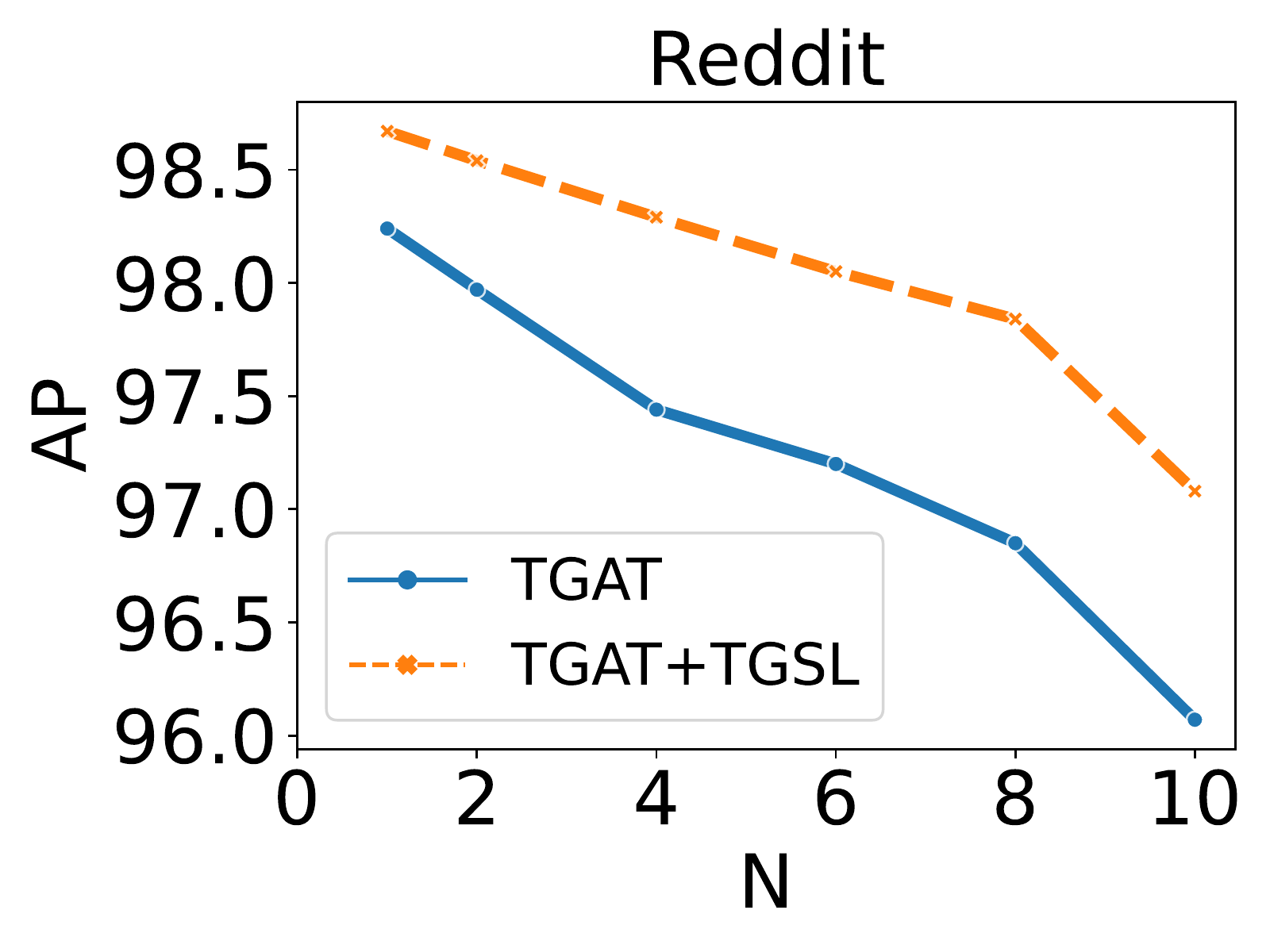}
		\caption{Transductive AP}
		\label{reddit_ap_trans}
	\end{subfigure}
	\centering
	\begin{subfigure}{0.24\linewidth}
		\centering
		\includegraphics[width=1.0\linewidth]{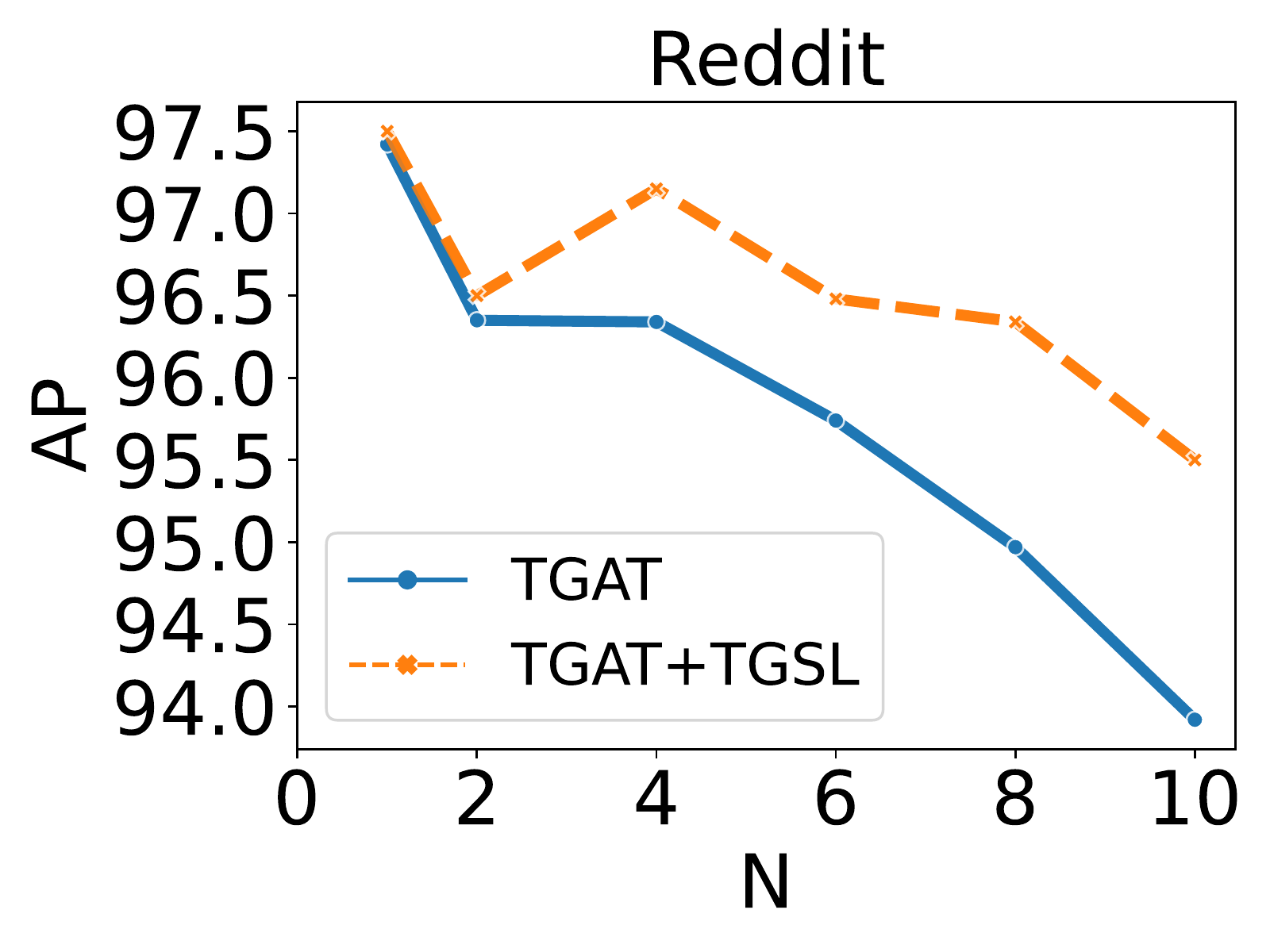}
		\caption{Inductive AP}
		\label{reddit_ap_induc}
	\end{subfigure}
	\caption{Performance on Graphs with Different Sparsity. We reserve one edge for every $N$ interactions on the Wikipedia and Reddit datasets. $N=1$ represents no interaction deletion.}
% 	\vspace{-1ex}
	\label{scarcity}
\end{figure*}

To investigate how TGSL performs on graphs with different sparsity, we artificially remove a varying number of interactions from both the Wikipedia and Reddit datasets. 
Specially, we reserve one edge for every $N$ interactions, while keeping the validation and test edges unchanged. 
We then retrain the TGAT and TGSL on the modified datasets to verify whether TGSL can still maintain a competitive performance in the case of severe edge missing. 
The results, as measured by AP under both transductive and inductive settings, are presented in Figure \ref{scarcity}. 
We observe that both TGAT and TGAT+TGSL experience a significant performance drop in AP under both transductive and inductive settings as the number of missing edges increases on the two datasets. 
Nevertheless, TGAT+TGSL still achieves a remarkably competitive result, consistently surpassing a single TGAT. 
Furthermore, as $N$ increases, the performance gap between TGAT and TGAT+TGSL widens. 
While a single TGAT shows a more drastic AP decline, TGSL is more moderate, which demonstrates that TGSL is relatively robust to the dataset with severely missing edges and is particularly beneficial for sparse datasets.

\subsubsection{{The Convergence Speed}}
\label{sec:convergence_analysis}

\begin{figure}[!ht]
% \vspace{-1ex}
	\centering
	\includegraphics[width=0.99\linewidth]{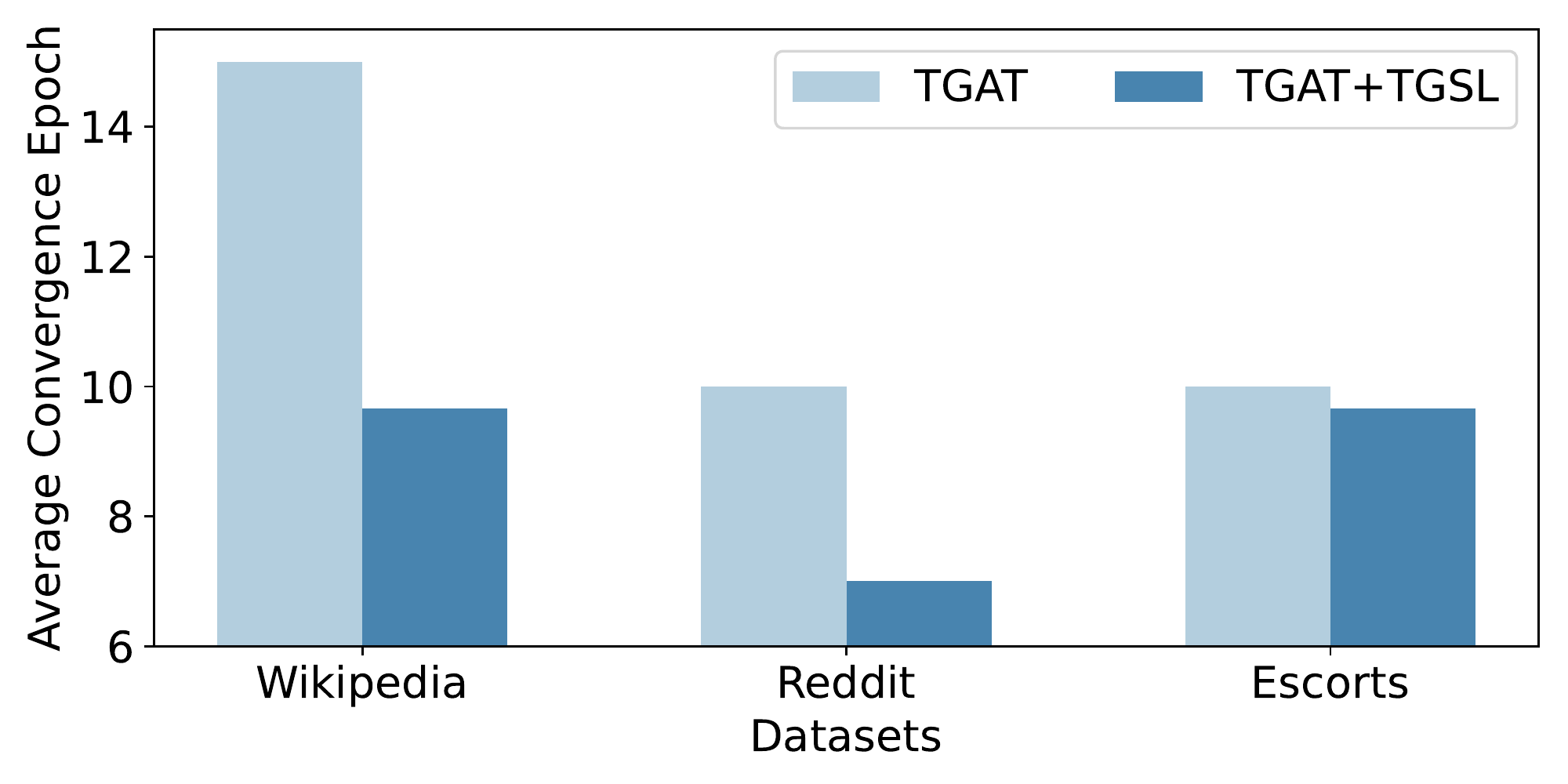}
% 	\vspace{-1ex}
	\caption{Average Number of Epochs for Convergence over 3 Runs of TGAT and TGSL. }
% 	\vspace{-1ex}
	\label{convergence}
\end{figure}

In this part, we analyze the convergence speed of TGSL on the Wikipedia, Reddit, and Escorts datasets. 
The hyper-parameters related to early stopping are detailed in Section \ref{exp:setup} and are kept consistent throughout all experiments. 
As shown in Figure \ref{convergence}, we present the average number of epochs for convergence on TGAT over 3 runs before and after integrating our proposed TGSL. 
Obviously, TGSL speeds up the convergence of TGAT on the three datasets. 
Especially, TGSL significantly accelerated the convergence of TGAT on the Wikipedia and Reddit datasets, which benefits from the TGSL's effective optimization of the graph structure, allowing TGAT to learn better graph representations more efficiently.

\subsubsection{{The Impact of the Number of Added Edges $K$ in TGSL}}
\label{sec:xxx}

\begin{figure}[h]
% \vspace{-1ex}
    \centering
	\begin{subfigure}{0.49\linewidth}
		\centering
		\includegraphics[width=1.0\linewidth]{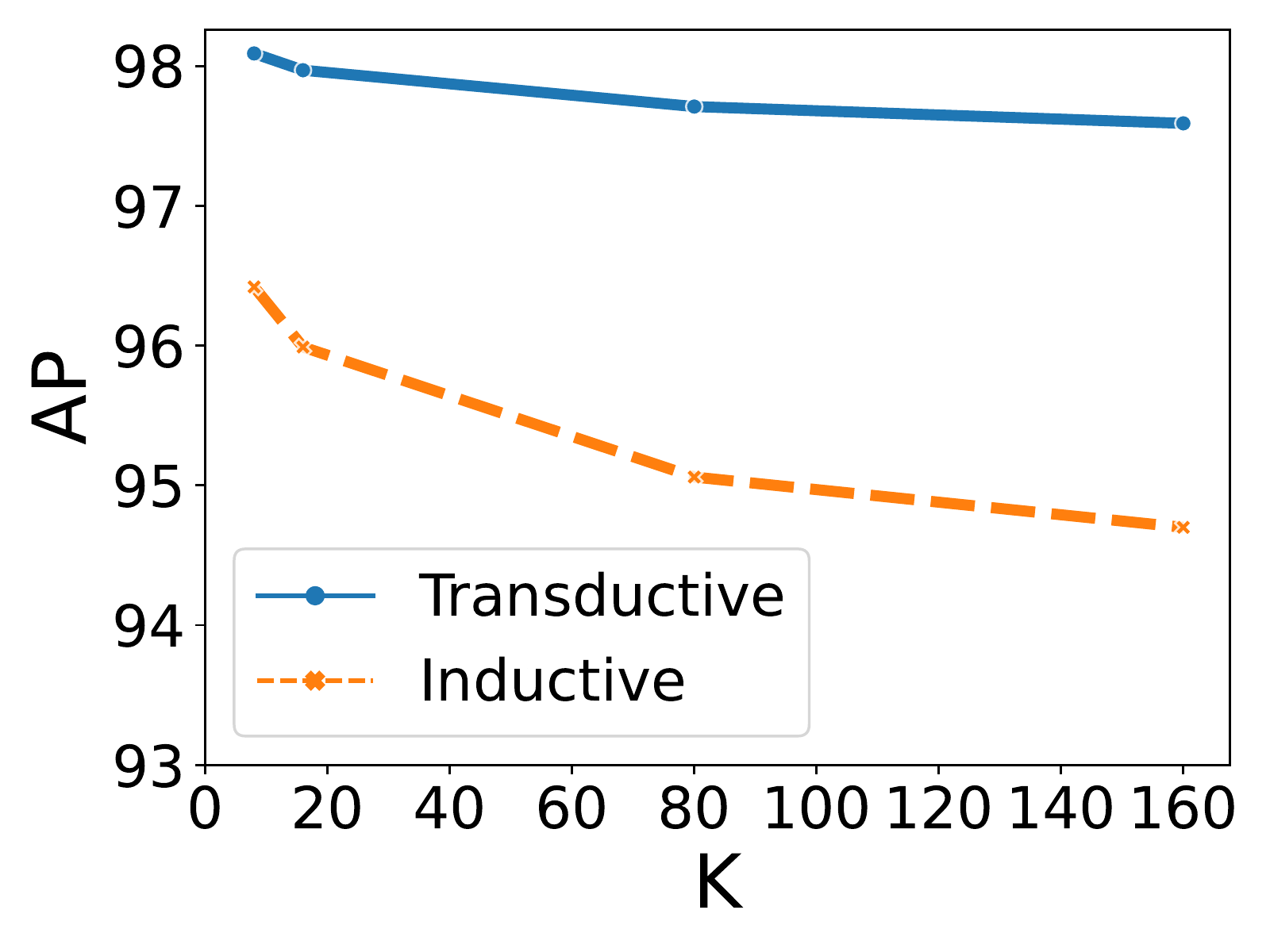}
		\caption{TGAT + TGSL}
		\label{tgat_k}
	\end{subfigure}
	\centering
	\begin{subfigure}{0.49\linewidth}
		\centering
		\includegraphics[width=1.0\linewidth]{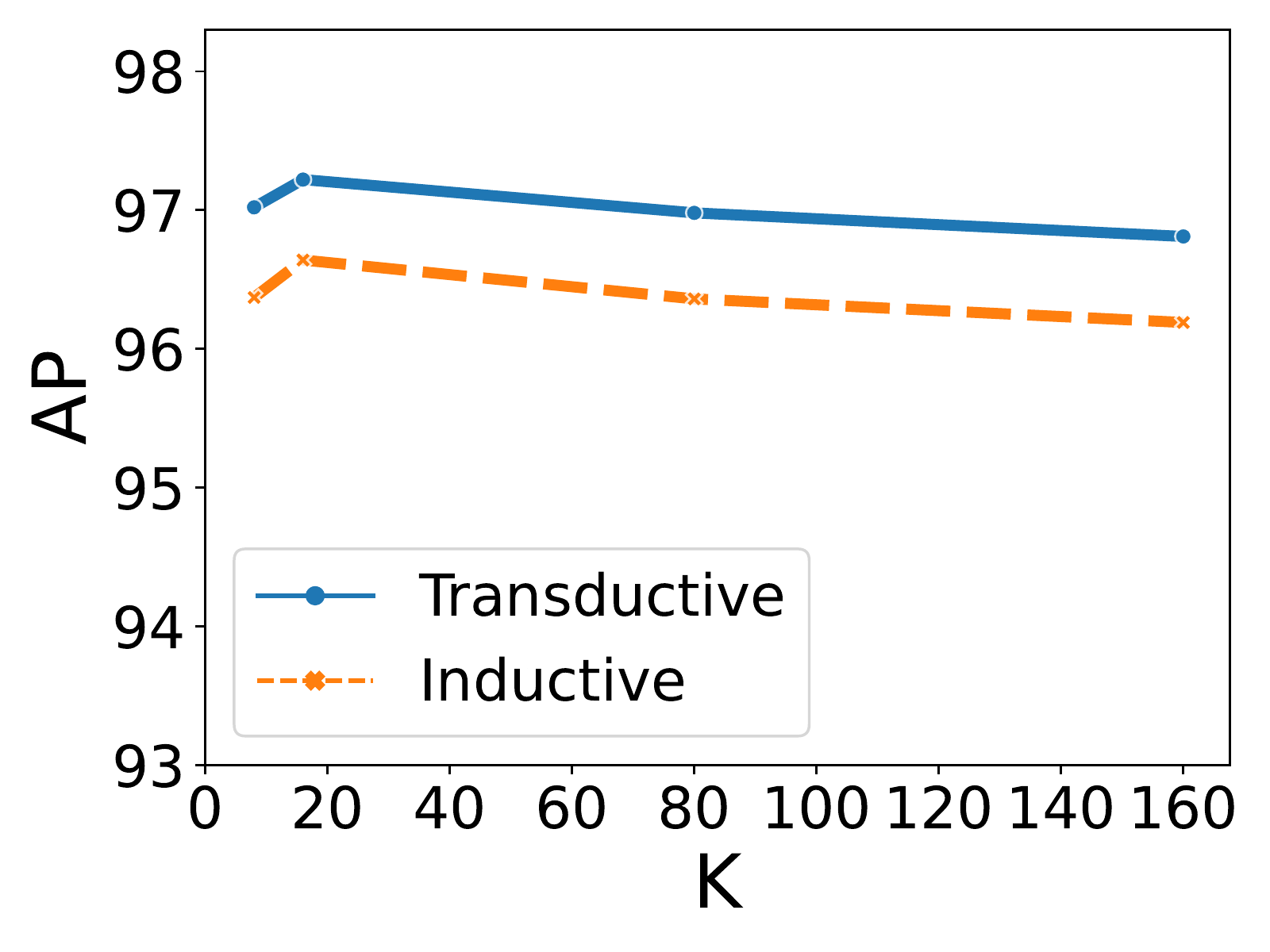}
		\caption{GraphMixer + TGSL}
		\label{graphmixer_k}
	\end{subfigure}
	\caption{Performance in Terms of the Number of Added Edges $K$ in TGSL on the Wikipedia Dataset. }
% 	\vspace{-1ex}
	\label{sen_k}
\end{figure}

In order to explore the influence of the number of added edges $K$ in TGSL, we use both TGAT and GraphMixer to conduct sensitivity analysis on the Wikipedia dataset under both transductive and inductive settings. 
In this part, the candidate edge sampling strategy is all set to 3rd-hop neighbor sampling, and we only change the number of $K$ while keeping the rest of the hyper-parameters unchanged. 
The AP results are shown in Figure \ref{sen_k}. 
From Figure \ref{sen_k}, we observe that a small $K$ is sufficient for TGSL to learn a better graph structure. 
Both TGAT and GraphMixer reach the best AP with a relatively small $K$ (i.e., $K=8$ and $K=16$, respectively) under the 3rd-hop neighbor sampling strategy. 
Both TGAT and GraphMixer experience performance drop as $K$ becomes larger, which demonstrates that adding more edges does not bring continuous improvement to TGSL. 
On the contrary, it may bring a lot of noise to the model and hinder the learning process of TGSL. 
Moreover, a large number of new edges can increase the computational burden on TGNs, making training inefficient.

\section{Conclusion}
\label{sec:conclusion}

The effectiveness of TGNs is heavily dependent on the quality of the graph structure. In reality, graph structures are often incomplete and noisy, which hinders TGNs from learning informative representations.
We propose a time-aware graph structure learning method to learn a better graph structure on temporal graphs. In particular, it predicts time-aware context embeddings based on previously observed interactions. Using these context embeddings, we can predict the nodes that are likely to interact at a given timestamp and construct the augmented graph. Additionally, we optimize TGNs and TGSL by employing supervised losses on the original and learned graphs, along with a contrastive loss between the two views, and perform inference on the refined graph.
Extensive experiments conducted on three public datasets demonstrate the effectiveness of TGSL. 
Our method currently focuses on refining incomplete graph structures through adaptive edge addition, but it can be easily extended to edge dropping for removing noisy edges. We leave this as future work.

\newpage

\bibliographystyle{ACM-Reference-Format}
\balance
\bibliography{ref}

\end{document}